\documentclass[letterpaper, 10 pt, conference]{ieeeconf}  % Comment this line out if you need a4paper

\IEEEoverridecommandlockouts 

\usepackage{graphicx}
\usepackage{mathtools}
\usepackage{comment}
\usepackage[bottom]{footmisc}
\usepackage{cite}
\usepackage{stfloats}
\usepackage{multirow}
\usepackage{subfigure}
\usepackage{algorithm} 
\usepackage{algpseudocode}
\usepackage{hyperref}
\title{\LARGE \bf
DeepScanner: a Robotic System for Automated 2D Object Dataset Collection with Annotations
}
\author{\authorblockN{Valery Ilin, Ivan Kalinov, Pavel Karpyshev, and Dzmitry Tsetserukou}
\authorblockA{ \textit{Skolkovo Institute of Science and Technology, Moscow, Russia 121205}\\
\{valery.ilin, i.kalinov, pavel.karpyshev, d.tsetserukou\}@skoltech.ru}}

\begin{document}

\IEEEoverridecommandlockouts

\pubid{\makebox[\columnwidth]{978-1-7281-2989-1/21/\$31.00 \copyright2021 IEEE\hfill} \hspace{\columnsep}\makebox[\columnwidth]{ }}

\maketitle
%\thispagestyle{empty}
%\pagestyle{empty}

%%%%%%%%%%%%%%%%%%%%%%%%%%%%%%%%%%%%%%%%%%%%%%%%%%%%%%%%%%%%%%%%%%%%%%%%%%%%%%%%
\begin{abstract}
In the proposed study, we describe the possibility of automated dataset collection using an articulated robot. The proposed technology reduces the number of pixel errors on a polygonal dataset and the time spent on manual labeling of 2D objects. The paper describes a novel automatic dataset collection and annotation system, and compares  the results of automated and manual dataset labeling. Our approach increases the speed of data labeling 240-fold, and improves the accuracy compared to manual labeling 13-fold. We also present a comparison of metrics for training a neural network on a manually annotated and an automatically collected dataset.

\end{abstract}

%%%%%%%%%%%%%%%%%%%%%%%%%%%%%%%%%%%%%%%%%%%%%%%%%%%%%%%%%%%%%%%%%%%%%%%%%%%%%%%%
\section{Introduction}

\subsection{Motivation}
The task of manual image segmentation in order to label each pixel corresponding to the target objects is significantly different from the tasks of classification and detection. To label one object in the image, it is necessary to add a huge number of control points to create a polygon of the labeled object. The most popular datasets, for example COCO \cite {lin2014microsoftCOCO} and PASCAL VOC \cite {everingham2015pascalVOC}, were formed in a similar way. On the basis of these datasets, a huge number of competitions are held in the areas of object detection and segmentation using computer vision and machine learning algorithms. Another example of such datasets is the IceVisionSet dataset \cite{pavlov2019icevisionset} collected using various sensors for autonomous vehicle development. Visual data annotation included road sign information, and, based on this dataset, a contest was performed for robust road sign detection in complicated weather and light conditions. % \cite{pavlov2019recognition}.
However, like most other datasets, these datasets were labeled manually, which naturally led to numerous pixel errors. 

% The manual markup of custom datasets occurs in a similar way.

\begin{figure}[htbp]
\centerline{\includegraphics[scale = 0.18]{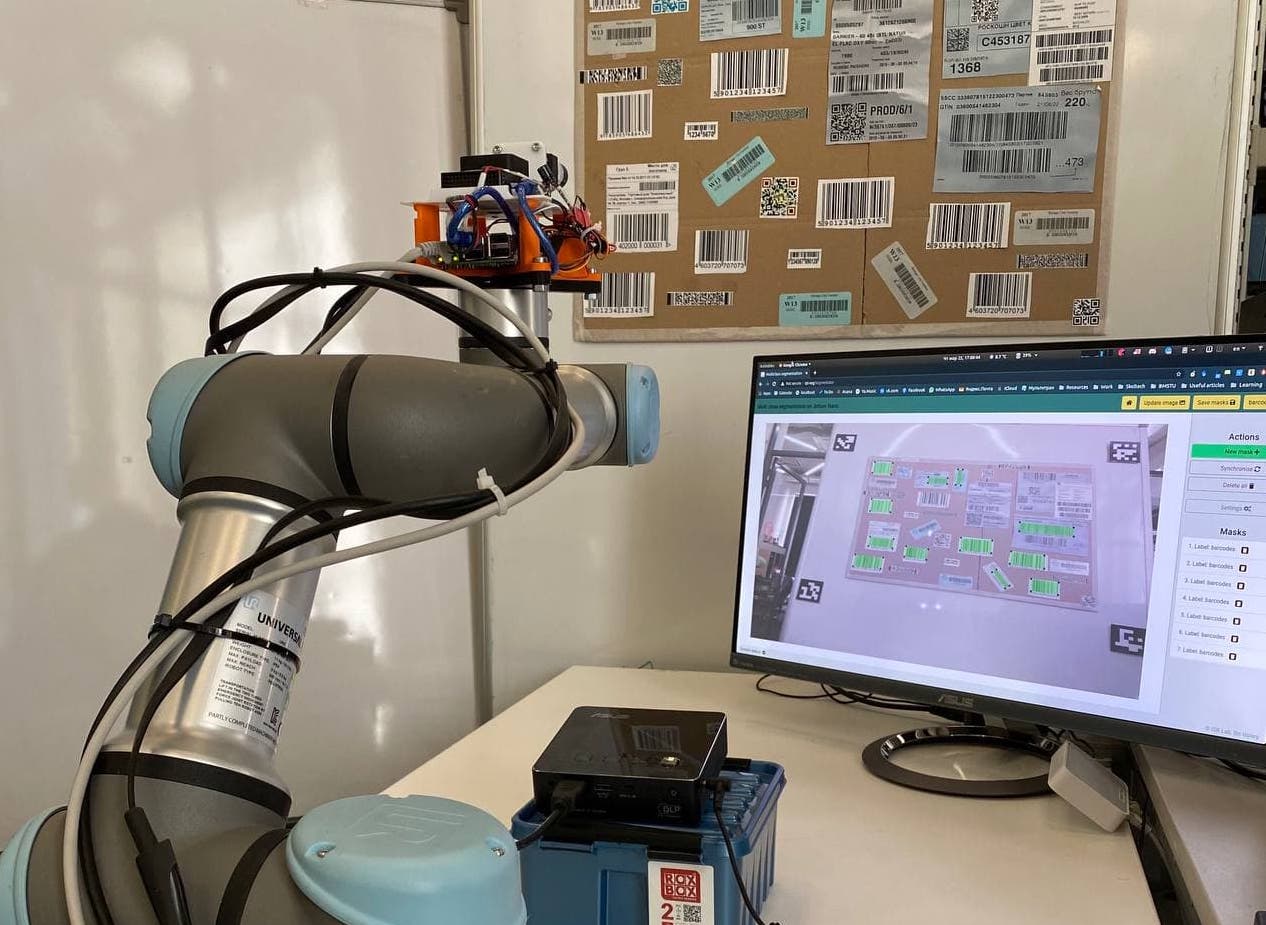}}
\caption{DeepScanner for automated data labeling system. Robot is facilitated with a set of sensors, a stand with targets for labeling, a data augmentation projector, and a graphical labeling system interface.}
\label{fig:main}
\vspace{-1.5em}
\end{figure}

The paper \cite{kalinov2020warevision} demonstrates the possibility to recognize barcodes in an image using a convolutional neural network (CNN) based on the U-net architecture \cite{ronneberger2015unet}. It revealed the high efficiency in object recognition, however the approach still results in a $\sim$ 9\% pixel error on segmentation tasks. First of all, this is due to independently collected and labeled data. If examined in detail, the dataset lacks several polygons in the image and contains small displacements of the polygon anchor points.
% In the existing system \cite{kalinov2020warevision}, we used a set of sensors located on the UAV. We used the Pixhawk micro-controller to get the exact position of the UAV and control it. As a source of additional positional data, these microcontrollers incorporate an Inertial Measurement Unit (IMU) to capture accelerations and movements indirectly. To capture the image, we used the Raspberry Pi NoIR camera v.2. Since this camera captures images using a rolling shutter, that is, the image capture from the matrix occurs line by line, then with a slight vibration of the device, image distortions may occur. To eliminate them, the system contains vibration dampers.  Finally, we used the Zebra Barcode scanner to scan 1D barcodes in the warehouse.
% All calculations were performed on-platform using a Jetson Nano (2019) microcomputer. This allowed us to recognize barcodes and obtain masks using a convolutional neural network in real time. 

%Automated dataset collection may also be useful in various areas like retail and agriculture. For example, high quality annotated data is crucial for autonomous plant disease detection, as stated in \cite{karpyshev2021autonomous}. As for retail, the results of the paper by Petrovsky et al. \cite{petrovsky2020customer} , \cite{yatskin2017principles}could be further extended to products without RFID markers on them, using only the visual and location data.

Automated dataset collection may also be useful in any task that involves models trained on annotated data, from searching for people by a group of drones \cite{yatskin2017principles} to recognition of goods on store shelves \cite{petrovsky2020customer}. For some tasks, high-quality annotated data is a key parameter for success, e.g., it is crucial for autonomous plant disease detection, as stated in \cite{karpyshev2021autonomous}, and for people detection during autonomous disinfection \cite{ultrabot}.

\subsection{Problem statement}
Despite almost all datasets nowadays are annotated manually, the approach is not accurate and robust enough.
The disadvantages of manual image labeling are discussed in the work of Chen et al. \cite{cheng2018surveyAIA} and they are still relevant to this day. The methods for automating image labeling presented in his work are based on mathematical and computer models, which may not always be applicable in real life on a large amount of similar data.

The problem of data validity in manual labeling is also described in Song et al. \cite{song2020validations}, where it is stated that manual coding of information is directly dependent on the expertise of the person labeling the data.

\begin{figure}
    \centerline{\includegraphics[width = 0.4\textwidth]{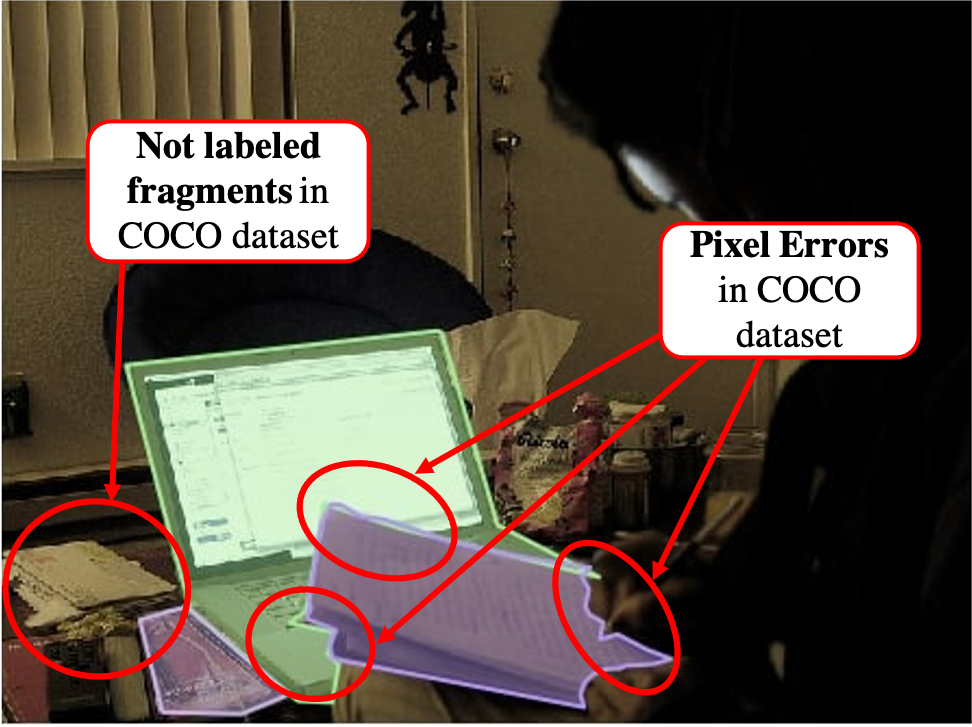}}
    \caption{Examples of errors in manual image segmentation from the COCO dataset (val2017) on image with ID 35279. The same issues were mentioned in images with ID 176446, 147415, and 348481.}
    \label{fig:coco_problems}
    \vspace{-1.5em}
\end{figure}

Recently, the WIRED magazine published an article \cite{wiredImageNet} describing errors in one of the ImageNet fundamental datasets \cite{deng2009imagenet}. It was estimated that the dataset contains approximately 6\% of errors, which is a fairly large volume of the total number of images (14 million labeled images). The reason for this is also manual data labeling: the data is typically collected and labeled by low-paid inexperienced workers.

To artificially increase a dataset of images, data augmentation is typically used. The most popular augmentation methods (basic image manipulations) are described in the work \cite {shorten2019survey}. These approaches can be performed at the stage of image preparation for subsequent transfer to the neural network input. In \cite {ma2019optimizingAUGM}, augmentation methods for image segmentation semantics are presented. When applying geometric augmentation (i.e., rotation, resizing, and cropping) the accuracy of the final model increases. However, it should be noted that the initial annotation pixel error remains with any of the methods of geometric augmentation of images and masks.

To solve the problem of accurate segmentation labeling of 2D objects (barcodes, QR codes, price tags, and etc.) in the image, we approach it based on the use of a 6 Degree of Freedom (6-DoF) robot with built-in precise position sensors to accurately track the coordinates of the camera (see Fig. \ref{fig:main}). DeepScaner aims not only to automate the dataset collection and annotation but also to decrease the numerous pixel errors presented in Fig. \ref{fig:coco_problems}. A detailed system description is presented in \autoref{methodology}.

\subsection{Related works}
\label{sec:related}

Automated data collection is a well-known approach and is actively used in modern research. In general, there are two directions on which this kind of research is based: analytical, based on sensors, and evaluative, based on machine learning models. The work of Ke et al. \cite {ke2019end} is based on the second approach and uses a pretrained CNN to generate textual descriptions of images (what classes of objects are in the image). This approach has shown its effectiveness in comparison with manual data labeling and also surpassed existing approaches on automated image labeling with text. However, errors remain in the labeled data: precision of the approach directly depends on the training data, which, in turn, also has errors. 

The work of Song et al. \cite{song2020weighted} also describes automated image annotation. Their approach includes the use of an intermediate layer in a neural network to extract data from images more accurately. The approach allows making accurate textual descriptions of images, however, it is not versatile: when new data appears, it is necessary to retrain the model on a new dataset, which must be labeled manually.
More accurate data annotation is provided by approaches based on the operation of accurate sensors for obtaining information and analyzing their working principles.

Jenson et al. \cite{jensen2014large} presented an MVS dataset assembled using a 6-DoF robot and a set of depth cameras. The dataset is quite accurate, since in its construction, position sensors and data from cameras were used to plot the depth of objects.

The work by Ruiz-Sarmiento et al. \cite{ruiz2015olt} describes an automated labeling system based on a point cloud from an RGB-D camera. Their approach uses boxes to label the first frames. The resulting box labels are anchored in space and subsequent captured frames are labeled based on the position of the camera in space. The accuracy of this approach directly depends on the accuracy of the algorithm for determining the RGB-D camera position.

Stumpf et al. \cite{stumpf2021salt} describe the layout of images with 3D bounding boxes using an RGB-D camera. Images were labeled using the information about the current depth-map of the frame.
% on depth analysis, selection and objects. 
This approach is useful for labeling people or other moving objects, however, the fixed camera angle does not create additional augmentation of the background data. Additionally, it requires control of the labeling system by the operator. Their labeling method showed impressive results in terms of speed and labeling accuracy. Authors managed to increase the labeling speed 33.95 times for building a bounded box object and 8.55 times for a segmentation task compared to manual dataset labeling.

Automated dataset labeling is also important for areas such as autopilot. Martirena et al. \cite{martirena2020automated} described a new method for automated annotation of road lane labeling. The approach relies on point cloud data and vehicle odometry. With the help of an automated annotation algorithm, the data is processed in blocks following the trajectory of the vehicle. As a result of the work, automatic labeling of road lanes is obtained, which in terms of labeling time is 60\% faster than manual data labeling. Moreover, the construction of segments does not exceed an error of more than 8\% with an error of 5 cm, which is a very good result for a similar task if performed manually.

De et al. \cite{de2019semiautomatic} describe the use of a 6-DoF robot and ArUco tags to create semi-automatic annotation of real-life objects (fruits, vegetables, and etc.). Their approach made it possible to reduce the time for labeling one frame 350-fold, while the labeling accuracy increased by 15\%. However, their approach uses bounding boxes for annotation, which is not suitable for object segmentation tasks.

All works reviewed in this study describe the automation of data collection and labeling in order to increase accuracy and reduce the time spent on data labeling. However, none of the works consider automated labeling of 2D objects. This task can be critical in cases where it is necessary to recognize similar objects and divide them into required classes.

\subsection{Contribution}

To solve the problem of automated labeling of 2D objects, e.g., barcodes, QR codes, price tags or other flat objects on the image, we developed DeepScanner, an automated labeling system using a 6-DoF robot with an end-effector containing a camera and auxiliary sensors for obtaining the end-effector position. Preliminary, the operator needs to annotate one frame in the developed data labeling system, thus, all the reference polygons of the labeled objects will be obtained. After activating the labeling algorithm, the industrial robot will perform random movements in a given area. Simultaneously, the accurate end-effector position is obtained, after which the points of initial matrices are transformed.

The proposed approach allows to:  
\begin{itemize}
    \item significantly increase the sample size. Previously, the manually assembled dataset consisted of 650 images, with 7-8 objects on average. More than 50 hours of manual labeling were spent on labeling whole dataset. The proposed collection system allows to label more than 30 frames with 10-15 objects on each in less than 1 minute;
    \item significantly increase the accuracy of polygon labeling. The conducted experiments have shown that image labeling using the proposed system is 13 times more accurate compared to manual labeling. At the same time, the main focus of the comparison was on the pixel-level accuracy;
    \item apply non-synthetic augmentation on colors and positions.  Instead of fixed lighting, a projector focused on the object board was used, lighting the objects with randomly generated colors. This approach made it possible to give the board with target objects various shades of colors that imitate the real lighting in different situations (e.g., various light conditions in industrial warehouses).
 \end{itemize}

\section{System Overview}

\subsection{Hardware layout}
\begin{figure*} [!t]
    \begin{center}
        \vspace{-1.5em}
        \includegraphics[width=0.9\textwidth]{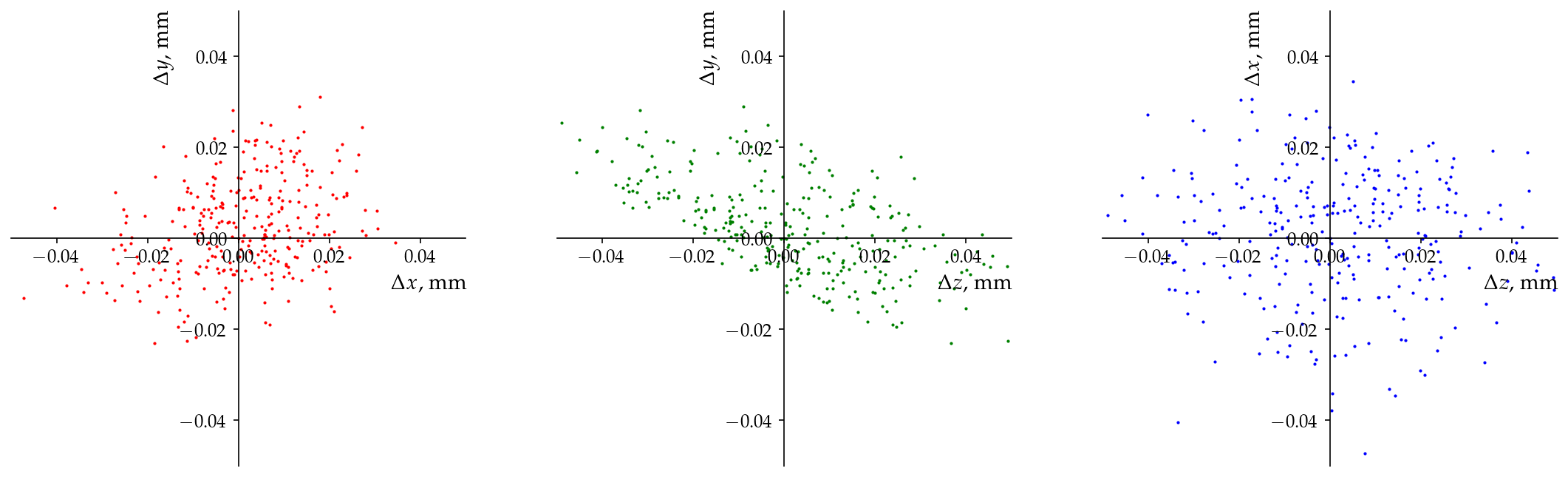}
        \vspace{-1.5em}
        \caption{Repeatability deviation of the 6-DoF robot along the $x$, $y$, $z$ axes.}
        \label{fig:delta_6dof}
        \vspace{-1em}
    \end{center}
    \vspace{-1em}
\end{figure*}

The proposed system is based on the 6-DoF UR3 collaborative robot. It is used to automate the process of moving the end-effector with the camera module to capture a set of frames, and to record the accurate end-effector position at the time of frame capturing.
According to the specifications of UR3, position setting accuracy of robot end-effector is 0.1 mm (repeatability parameter). According to the real experiments \cite{urRepeatability} the repeatability of UR3 is about 0.01 mm. We conducted our own measurements to confirm the measurement accuracy of this robot. To do this, the joins of the robot were iteratively returned to the original position after each random move and data on the position of the end-effector was recorded. As a result, the robot returned to its original position 300 times, and, out of 300 measurements, the absolute value of the offset by the x-axis did not exceed 0.045 mm, by the $y$-axis 0.032 mm and on the $z$-axis 0.05 mm. The graph of offsets along three axes is shown in Fig. \ref{fig:delta_6dof}. The error for the rotation parameter on the $x$, $y$, and $z$ axes did not exceed 0.08 degrees. Thus, the precise position of robot end-effector is known at each moment. These measurements were chosen as a ground truth parameter for the proposed labeling system. All experimental sensors were installed on the robot end-effector.

\begin{figure}[!t]
    \centerline{\includegraphics[width=0.45\textwidth]{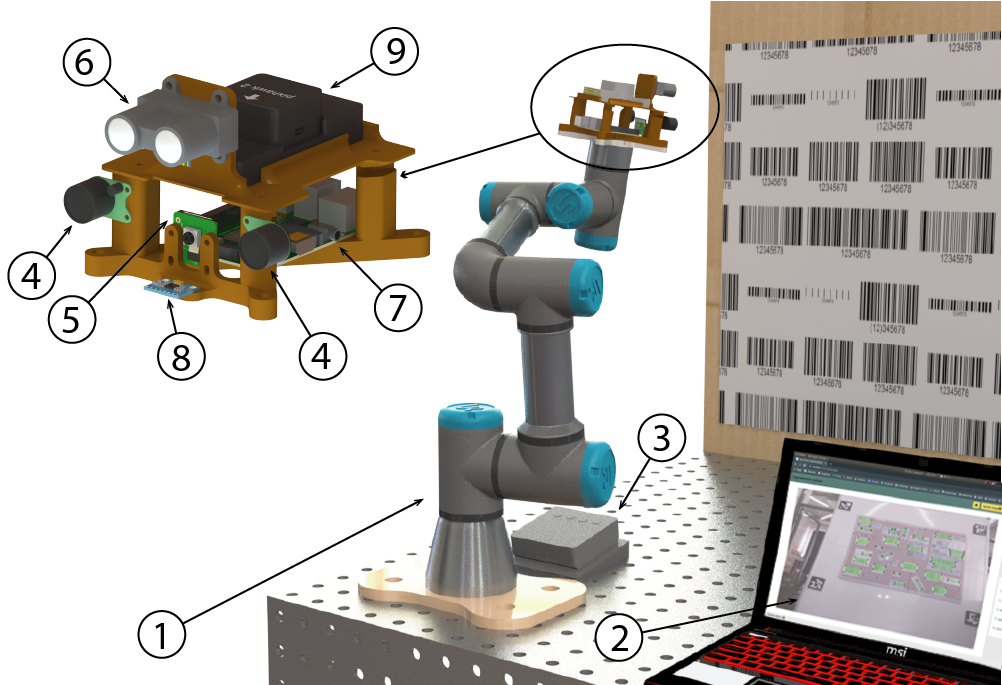}}
    \caption{Hardware system overview. 1: 6-DoF collaborative robot, 2: segmentation software system (operator console), 3: projector for light augmentation, 4: IR LEDs, 5: Raspberry Pi camera v.2, 6: Garmin LIDAR-Lite v3 sensor, 7: Raspberry Pi 3 microcomputer, 8: IMU sensor, and 9: Pixhawk flight controller.}
    \label{fig:harware-render}
    \vspace{-1em}
\end{figure}

In order to achieve automated dataset collection, the system uses a number of sensors. A Raspberry Pi NoIR v.2 CSI camera with the IMX219 sensor is used for image capturing allowing capturing IR spectrum as well. Double IR LEDs (1 W, 850 nm) with automatic brightness adjustment are used to simulate different lighting conditions in IR spectrum. The transformation matrix from robot end-effector to camera frame could be presented as follows:

\begin{equation} 
\resizebox{0.9\columnwidth}{!}{$
\begin{split}
&{}^{E}T_{C}=  % split start with &
\begin{bmatrix}
{}^{E}R_{C}  & {}^{E}P_{C} \\\hline
\begin{matrix}
0 & 0 & 0
\end{matrix} & 1 
\end{bmatrix}=\\  % split start with \\ &
&=\left[ \begin{array}{cccc}
1 & 0 & 0 & \Delta x \\
0 & \cos{\theta} & -\sin{\theta} & \Delta y \\
0 & \sin{\theta} & \cos{\theta} & \Delta z \\
0 & 0 & 0 & 1 \\
\end{array}\right] =
\left[ \begin{array}{cccc}
1 & 0 & 0 & \Delta x \\
0 & 0 & -1 & \Delta y \\
0 & 1 & 0 & \Delta z \\
0 & 0 & 0 & 1 \\
\end{array}\right],  
\end{split}
\label{eq:transformation_robot_camera}  
$}  
\end{equation}
where ${}^{E}T_{C}$, constant value, is the transformation for the camera matrix with respect to (w.r.t.) end-effector of collaborative robot;
${}^{E}R_{C}$ is the rotation matrix for the camera frame attached to the end-effector;
${}^{E}P_{C}$ is the translation of frame origin attached to the end-effector of collaborative robot;
$\theta$ is the end-effector rotation angle along $x$ axis to obtain camera matrix rotation matrix (in our case $\theta = \pi/2$);
$\Delta x$, $\Delta y$ and $\Delta z$ are the translation parameters, which were chosen from CAD scheme of experimental setup presented in Fig. \ref{fig:harware-render}.

Also, it is  necessary not only to know the $x$ and $y$ positions of 2D target objects from the constructed mask, but also to determine the depth at which this object is located. For this, a Garmin LIDAR-Lite v3 single-beam laser range finder  was used. The sensor has an accuracy of determining the distance of $\pm 2.5$ cm at a range about 5 m, while the distance to barcodes in our setup did not exceed 1.2 m.  

In order to avoid the depth error on initial position, 1000 range finder measurements are made immediately before starting the labeling system. These measurements allow us to calculate the mean value, as well as to find the measurement error.

\begin{figure*} [!b]
    \begin{center}
        \vspace{-1.5em}
        \includegraphics[width=0.8\textwidth]{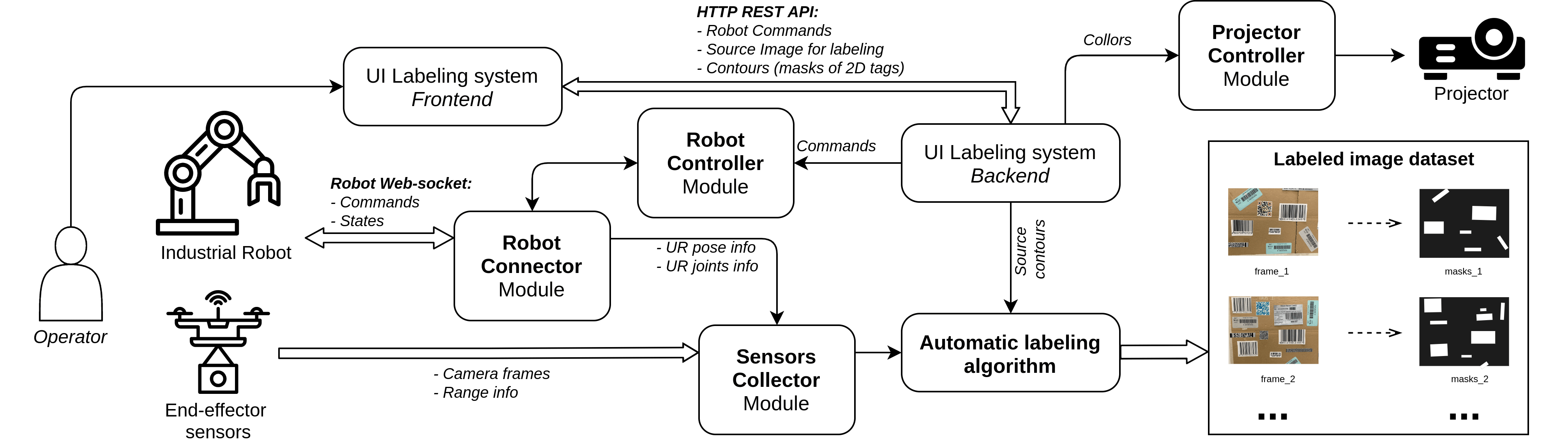}
        \vspace{-1em}
        \caption{The scheme of interaction of the software components and modules in the system of automated data labeling using an articulated robot.}
        \label{fig:software}
    \end{center}
\end{figure*}

In order to be able to use the obtained data for further improvement and debugging of the WareVision system \cite{kalinov2020warevision, kalinov2019high}, the Pixhawk flight controller and an IMU sensor were also installed on the end-effector. These sensors do not change the system operation principles and additional hardware can be installed based on the particular goals of the system.

A typical warehouse barcode is a set of black vertical lines on a flat surface presented in Code128, EAN-13 or other famous format. However, the background for the barcode lines are not necessarily white: brightly colored barcodes can often be seen in warehouses. Blue, green, and other contrasting colors can be used as background to attract the attention of labor worker during manual inventory. Warehouses may also have different lighting conditions, which may also affect the detection of  barcodes using the neural network.  To solve this problem, we decided to use the ASUS ZenBeam S2 projector to generate real-time color masks and white light with various color temperature (from cold blue to warm yellow). Thus, when training a neural network,  the data collected using the proposed system already contains additional color augmentation.

\subsection{Software architecture}

The software part of the labeling system is based on the developed algorithm for labeling objects (see Fig. \ref{fig:software}). To control the whole system, a frontend react-based web-site was developed. It is designed to visualize the current image from the camera, to obtain information from robot, to control robot position, to label frame with target objects, and to manage them. In order to transfer information between different components, a Flask web framework-based backend that connects with the frontend component was developed. Communication between all components is carried out using ROS topics. To communicate with the articulated robot, the HTTP 1.1 protocol (Web-Socket) was used, located in the \textit{Robot Connector Module}. To control the UR3, the \textit{Robot Controller Module} was utilized. \textit{The Robot Connector Module} also sends information about the robot position 25 times per second. The information from the sensors installed on the end-effector is collected using the \textit{Sensors Collector Module}. The following ROS topics with data are used in the system:
\begin{itemize}
    \item /camera\_image [ROS Image Message];
    \item /range\_data [ROS Range Message];
    \item /ur\_pose [ROS Position Message];
    \item /ur\_joints [ROS Float32MultiArray Message];
    \item /imu\_data [ROS Imu Message].
\end{itemize}

The collected information is transferred to the \textit{Automated Labeling Algorithm Module}, that outputs a folder with the labeled dataset: for each image file from the camera, a corresponding mask image and a file with the position of the contours are created. 

To make geometric transformations, the \textit{math3d} and \textit{numpy} python libraries are used. To generate masks and make visualization, we take advantage of the \textit{OpenCV} library. This library was also applied to generate color masks and output them onto the board with barcodes using a projector.

\section{Research Methodology}
\label{methodology}

The approach for the automated labeling of 2D objects in the image is based on the initial position of the end-effector, the image from the camera, and the current position of the end-effector. 

\begin{figure}[!t]
    \includegraphics[width=0.48\textwidth,clip]{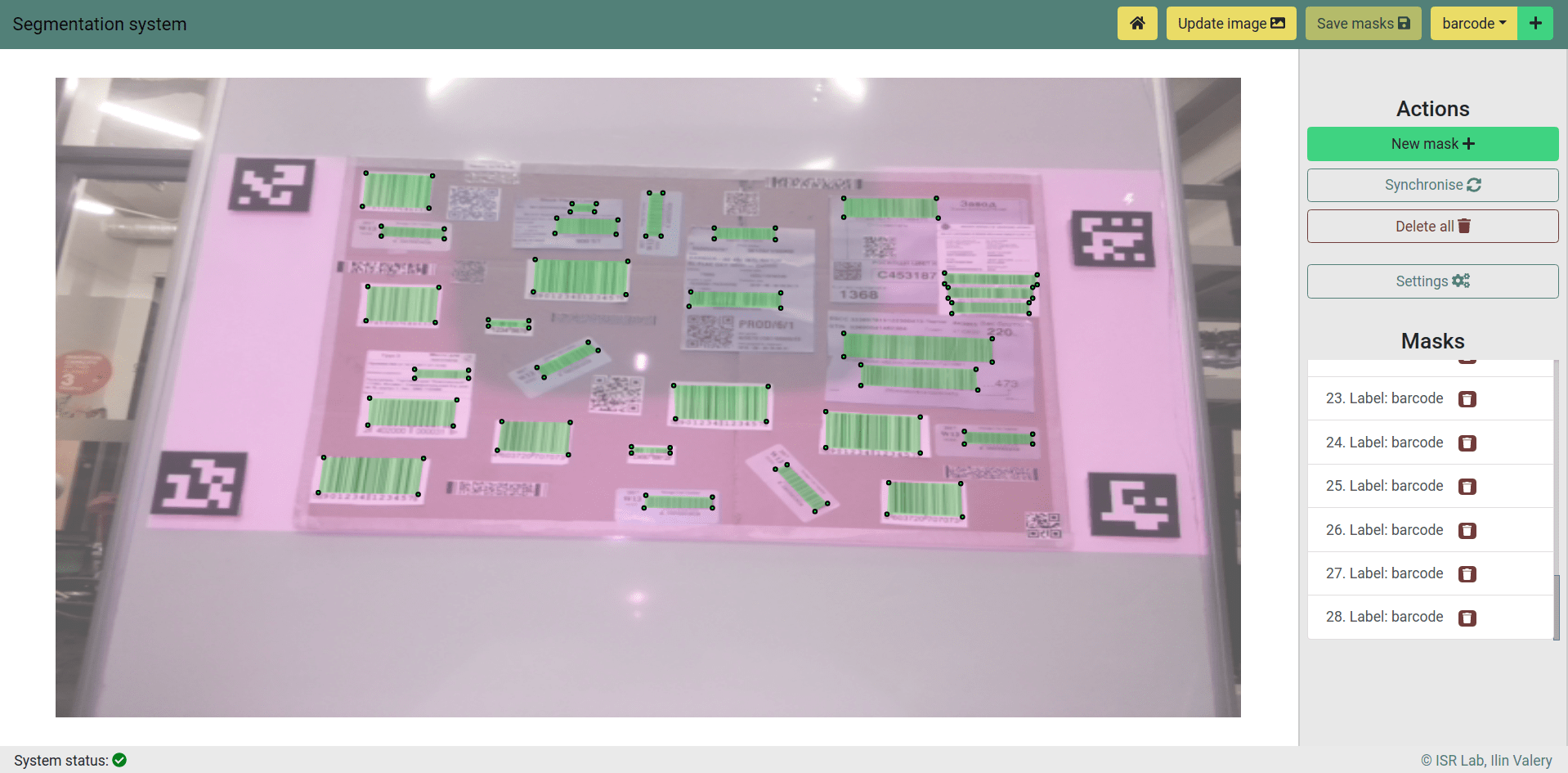}
    \centering
    %\vspace{-1em}
    \caption{User interface of the labeling system, where the barcodes are target objects for labeling.}
    \label{fig:UI_system}
    \vspace{-1em}
\end{figure}

First, it is necessary to prepare for the labeling. To do this, the original image from the camera installed on the articulated robot and the initial position of the end-effector ${}^{R}T_{E}$ in relation to the real world are collected in the User Interface (UI) component that is demonstrated in Fig. \ref{fig:UI_system}. The initial position of the surface of the camera matrix is parallel to the board with target objects. The next step is labeling the objects on the image. 
An example of the interface for image labeling is shown in Fig. \ref{fig:UI_system}. 
It is necessary to carefully label each required object. Though the experimental setup contained only objects of one class, the labeling system also supports multiple class annotation. After obtaining the image labels, a corresponding file will be created for each class, which contains the labeled polygon point coordinates in the image.

The initial parameters for our automated labeling approach are as follows:
\begin{itemize}
    \item ${}^{B}T^0_{E}$ is the initial transformation of end-effector w.r.t. the UR3 base. 
    \item $CM_{0}$ are the coordinates of masks on the initial frame for each class. In experiments, a single class is used. 
    \item $A$ is the camera calibration matrix;
    \item $D$ is the distance to the board with 2D objects.
\end{itemize}

Firstly, we obtain the transformation of camera w.r.t. UR3 base ${}^{B}T^0_{C}$ using parameter ${}^{B}T^0_{E}$:

\begin{equation}
    \label{eq:transform_initial}
    {}^{B}T^0_{C}= {}^{B}T^0_{E} * {}^{E}T_{C}.
\end{equation}
Next, we obtain a pose vector in the initial position of the camera  $P_{0} = (x_0,y_0,z_0,\alpha_0,\beta_0,\gamma_0)$.

The average value of distance $D$ is obtained immediately before the data labeling algorithm is run. To calculate the correct average value from the Garmin LIDAR-Lite v3 sensor, data accumulation and averaging are performed, which results in a confident accuracy in the obtained data. The achieved deviation $\sigma = 0.63$ cm  corresponds to the device specification and the obtained average was confirmed in several experiments.

\begin{figure}[!b]
    \vspace{-1em}
    \includegraphics[width=0.48\textwidth,clip]{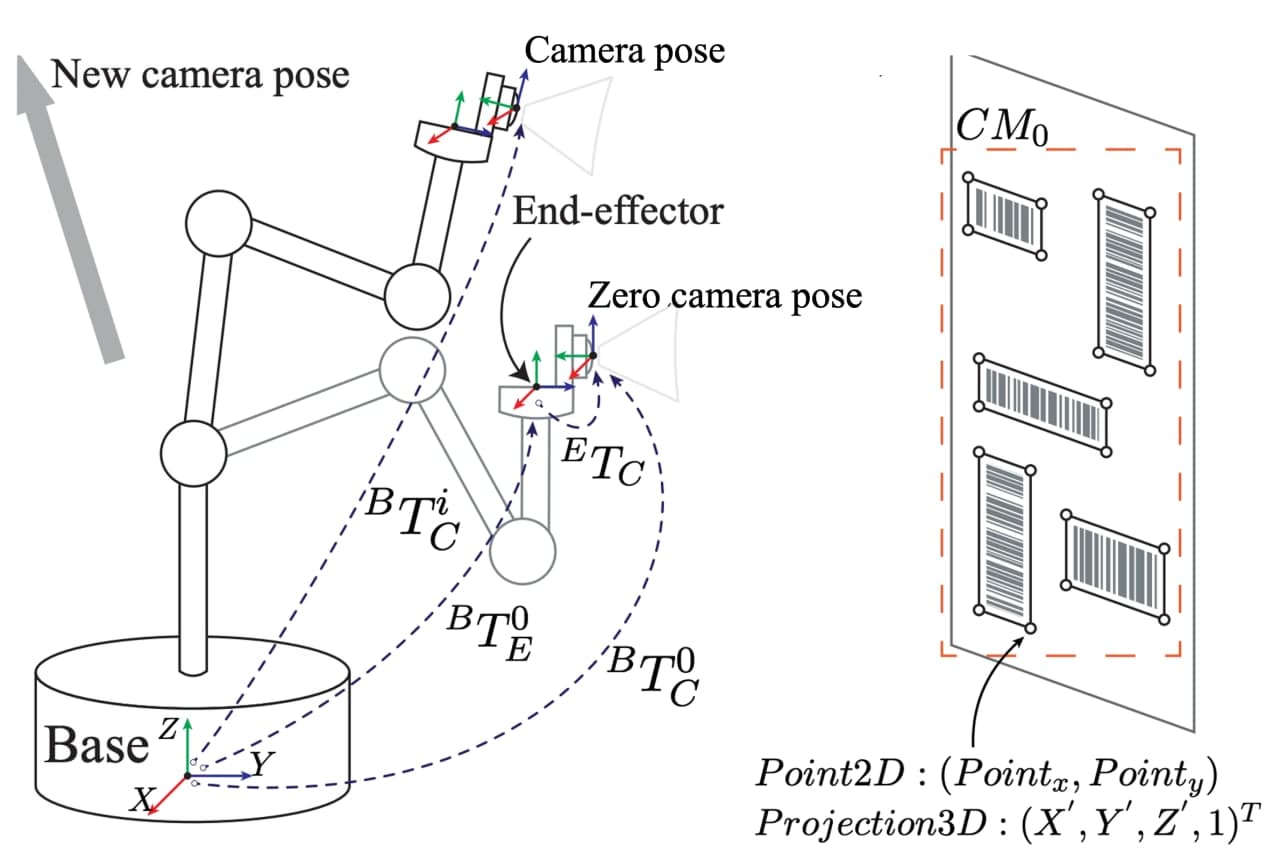}
    \centering
    \caption{Shifting the camera relative to the initial pose to the new camera pose $i$ and explaining the transformation matrices in the proposed approach.}
    \label{fig:methodology}
\end{figure}

Immediately after running the algorithm for automated data labeling, the system is ready to accept two objects for their subsequent storage and processing: the image from the camera and the current position of the 6-DoF robot. 
After acquiring the new position of the end-effector, the camera position is updated via the transformation matrix ${}^{B}T^i_{C}$:
\begin{equation}
    \label{eq:transform_i}
    {}^{B}T^i_{C}= {}^{B}T^i_{E} * {}^{E}T_{C}.
\end{equation}

Then, the resulting matrix is translated into the pose vector $P_{i} = (x_i,y_i,z_i,\alpha_i,\beta_i,\gamma_i)$. On the next step $\Delta P$ is calculated, that is the difference between initial and current poses:

\begin{equation}
    \label{eq:delta}
    \Delta P_i= P_{i} - P_{initial}.
\end{equation}

The processing of the initial labeled frame $CM_{initial}$ with current $\Delta P_i$  for each frame is demonstrated in Algorithm 1.

In the algorithm, the following parameters are used:
\begin{itemize}
    \item $F_x$, $F_y$ are the real focal distances given in the camera matrix;
    \item $w$, $h$ are the width and height of the frame;
    \item $scale$ is hhe proportional constant, depends on the distance from the camera matrix to center of the board with objects $D$; 
    \item $DC$ is the lens distortion coefficient.
\end{itemize}

The ${}^{B}M_{C}$ matrix [$3\times4$] is calculated as follows: 

% The demonstration of calculating the ${}^{B}M_{C}$ matrix [$3\times4$] is presented in \autoref{eq:camera_M}. 

\begin{equation} 
    \label{eq:camera_M} 
    {}^{B}T_{C} = 
        \begin{bmatrix}
            {}^{B}R_{C}                          & {}^{B}P_{C} \\\hline
            \begin{matrix} 0 & 0 & 0 \end{matrix} & 1 
        \end{bmatrix}= \begin{bmatrix}
            {}^{B}M_{C}  \\ \hline
            \begin{matrix} 0 & 0 & 0 & 1 \end{matrix} 
        \end{bmatrix}. 
\end{equation}

To get only two points ($x$ and $y$) from the final $Projection3D$ to make a corresponding mask, only two first matrices from the point projection coordinates are used:

\begin{equation} 
    Projection3D =  
    \begin{bmatrix}
        X^{'} \\
        Y^{'} \\
        Z^{'} \\
        1 \\
    \end{bmatrix}.
    \label{eq:projection3d}  
\end{equation}

The scheme describing the operation of the algorithm is shown in Fig. \ref{fig:methodology}.

The time spent on processing a single point in the algorithm does not exceed 40 ms. This is due to the use of small-dimensional matrix operations which are generally computationally light.

Then, after the final transformations, the resulting mask $CM_{frame_{id}}$ is saved in JSON format, and also converted to 2D polygons to save them as binary masks for the original images from the camera. This operation is performed using the OpenCV library, and pixels that are outside the scope of the original image size are cropped and ignored.

Thus, the system outputs a binary mask in the form of an image and a JSON file with the coordinates of the polygons for each image. Optionally, it is possible to output a combined image to compare the actual image with the labeling results.

The description of the accuracy  of this approach is presented in \autoref{seq:experiment1}.

\begin{algorithm}
\label{algo:main_transformation}
\fontsize{11pt}{11.5pt}\selectfont
	\caption{Automatic dataset collecting and labeling} 
	\begin{algorithmic}[1]
	    \State $CM_{frame_{id}}$ = $\emptyset$;
	    \For{$Mask$ in $CM_{0}$}
	        \State $NewMask$ = $\emptyset$;
            \For{$Point: (Point_x, Point_y)$ in $Mask$}
                \State $Point_{x\_real} $ = $Point_x * F_x / w - F_x / 2$;
                \State $Point_{y\_real} $ = $Point_y * F_y / h - F_y / 2$;
                \State $Point_{3D}$ = $[Point_{x\_real}, Point_{y\_real}, 1]^T$;
                \State ${}^{B}M^i_{C} = getCurrentRelatedPosition()$;
                
                \State $Projection3D$ = $A * {}^{B}M^i_{C} * Point_{3D}$;
                \State $Projection2D$ = $getProjection2D(DC)$;
                \State $NewMask$.append($Projection2D$);
            \EndFor
            
            \State $CM_{frame_{id}}$.append($NewMask$);
        \EndFor
    \State JSON.save($CM_{frame_{id}}$)
	\end{algorithmic}
\end{algorithm}
\vspace{-1em}

\section{Experiments}

\subsection{Experimental setup}

\begin{figure}[!t]
    \centerline{\includegraphics[width=0.45\textwidth]{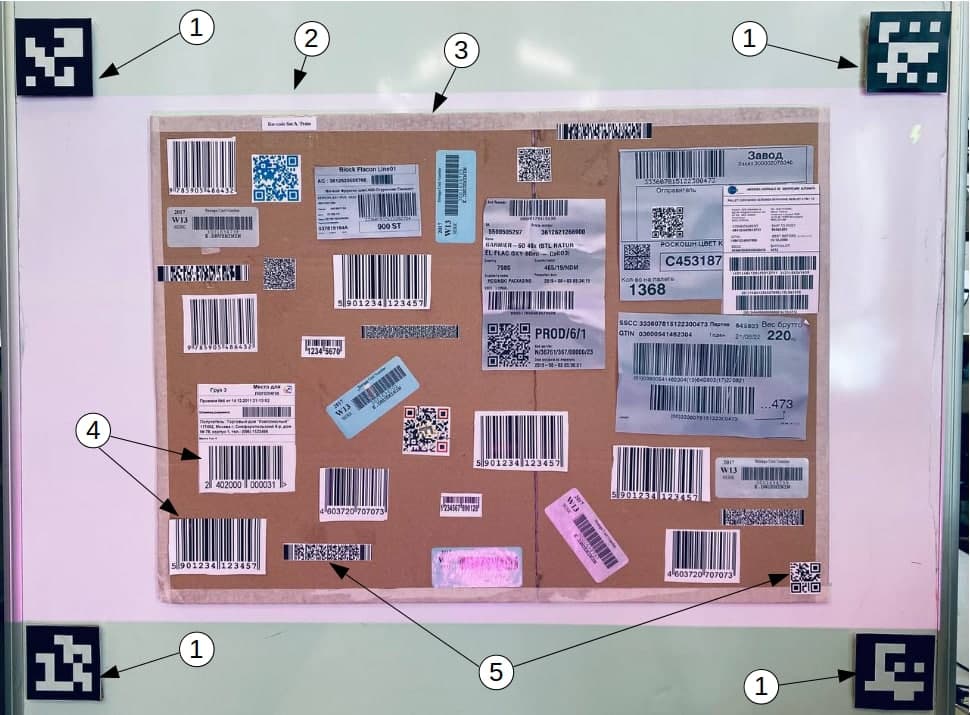}}
    \caption{Experimental field for labeling: 1 - ArUco markers, 2 - backlight from a short-focus projector, 3 - field with objects for labeling, 4 - examples of 1D barcodes (objects for labeling), 5 - examples of 2D barcodes and QR codes (not to be labeled).}
    \label{fig:stand-description}
    \vspace{-1em}
\end{figure}

For the experiments, a stand was developed (Fig. \ref{fig:stand-description}) containing a set of many types of 1D barcodes (only vertical lines). Also, to understand the labeling errors of objects (not pixel errors) during manual labeling, additional extra 2D barcodes and QR codes were added, which, according to the conditions of the experiment, do not need to be labeled.

Also, ArUco tags were added to the object board for experimental labeling without the use of an articulated robot. Using the tags, it is possible to evaluate the position of the camera relative to the experimental stand, and thus, to evaluate the new positions of the original masks using the proposed approach.

For image augmentation, the experimental stand is illuminated using a short-focus projector. At the output of the projector, a random light mask with different noises was generated, ensuring different illumination intensities and different colors of barcodes. The light in the room was turned off during the experiments.

The images obtained using the developed labeling system were transferred to operators for manual labeling. The labeling was done using one of the most popular online data labeling tools. For labeling, the experimental dataset was divided evenly between all participants of the manual labeling. Participants had to label all images (50 pieces per person) in the shortest possible time (without distractions) using the ``Polygon'' tool, which is setting points on the image in order to outline a specific object in the image.

\subsection{ArUco tags labeling approach}
This approach is not the main one in this paper, but it demonstrates an additional way of automated data labeling. The approach is based on the use of ArUco tags to estimate the position of the camera. 
Using ArUco labels was also described in section \autoref{sec:related} in the paper of De Gregorio et al. \cite{de2019semiautomatic}. ArUco tags do not require the use of expensive 6-DoF robots, so they can also be used for automated dataset labeling. However, the localization precision of this approach is significantly lower than the precision using the 6-DoF robot. Also, this method reduces the available area of camera movement: during experiments, the ArUco tags were often out of the camera FoV, thus, the localization was performed on 1, 2 or 3 out of 4 tags, which further reduced its accuracy.

\subsection{Experiment 1. Accuracy and speed evaluation}
\label{seq:experiment1}
The idea of the first experiment consists in comparing the indicators of time and labeling accuracy using three methods:
\begin{enumerate}
    \item manual labeling;
    \item automated labeling based on the position of ArUco tags;
    \item automated labeling based on the position of the end-effector of an industrial robot.
\end{enumerate}

As part of the experiment, 400 images of the stand were obtained from the device camera. 8 people, who had experience with the labeling system, took part in manual dataset labeling. Each person was asked to label 50 different images from the camera, with a time limit of 2 hours. Each person was asked to measure the time of data markup. As a result, the minimum time spent on annotating one frame was 38 seconds (on a frame with a minimum number of barcodes), the maximum was 127 seconds (full frame), the average for all participants of the experiment was 92.3 seconds. After the experiment, almost all subjects complained of slight fatigue. By the end of the data markup, the participants began to make more mistakes, which may indicate errors that occur during routine processes. 
After combining the data from all participants, a dataset of 400 manually labeled images was obtained.

To compare different approaches of  data labeling, the following 3 metrics were selected:
\begin{itemize}
    \item \textit{Mean frame time} is the time spent labeling a single frame. For automated labeling, the time spent on labeling the first frame is evenly distributed to all labeled frames. The time to label the original frame was 119 seconds, thus, when distributed over all 400 frames, the time for manual labeling of one frame with automatic data collection was 0.3 seconds. If the number of frames is increased, this indicator can be further reduced.
    \item \textit{Mean object error} is the \% of unlabeled or incorrectly placed objects in the image for the entire sample. To check this parameter, the overlap calculation was used. If the overlap of the labeled area relative to the original area was more than 75\%, the object was considered correctly labeled.

    \item \textit{Mean pixel error} is the number of incorrectly placed pixels. To obtain this metric, we randomly selected 20 frames, then labeled them up with high accuracy manually without a time limit and compared them with the usual manual labeling results.
\end{itemize}

Statistics for the three types of labeling are presented in Table \ref{tab:compare}. The error in the \textit{Mean object error} metric was eliminated on automated approaches due to the accurate initial frame labeling. The Auto ArUco method performed worse in the \textit{Mean pixel error} metric than the Manual labeling approach. This is due to the presence of frames where there were 1-2 ArUco tags in the frame, which adversely affects the exact position determination.

\begin{table}[!t]
\caption{Comparison of Methods for Labeling Barcodes in Images by Chosen Metrics}
\begin{center}
    \begin{tabular}{c|c|c|c}
                                 & Manual & Auto ArUco & Auto UR \\ \hline
    Mean frame time, sec &       92.3          &           0.42       &        0.34        \\
    Mean object error, \%  &        1.2     &           0          &      0              \\
    Mean pixel error, pixels  &     46.2        &       52.1                            &     3.3
    \end{tabular}
    
    \label{tab:compare}
    \end{center}
\end{table}

\subsection{Experiment 2. Model training}

\begin{figure}[h]
    %\vspace{-1em}
    \centerline{\includegraphics[width = 0.48\textwidth]{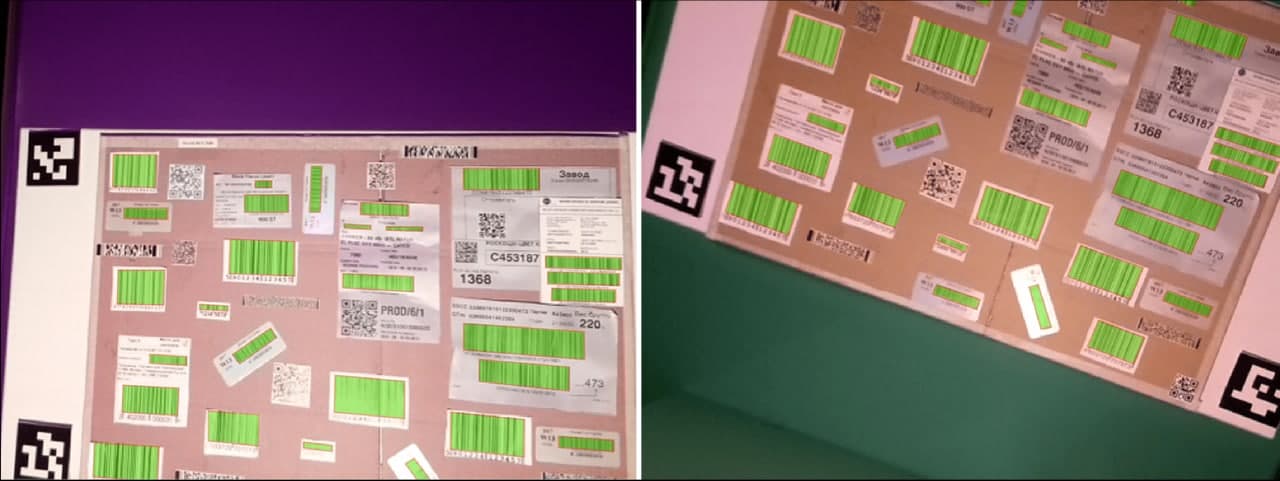}}
    \caption{Examples of image labeling using the proposed approach.}
    \label{fig:labeled_images}
    %\vspace{-0.5em}
\end{figure}

The second experiment is based on the use of labeled data obtained earlier to test the training of CNN. As the neural network architecture, we chose the previously described \cite {kalinov2020warevision} architecture based on U-Net with several modifications. As the input parameters of the network, an image of 640x480 pixels is used, in accordance with which all layers have been modified in terms of the size of the frame. Teaching parameters, the number of channels on each layer, activation functions, and other parameters remained unchanged. The neural network was trained for each of the proposed data collection methods. Each approach used the same original images with the appropriate labeling. Each dataset consisted of 400 images. The validation dataset consisted of 40 images; the test dataset had 40 images. Each image contained 20 barcode instances on average. An example of images is presented in Fig. \ref{fig:labeled_images}. For the training of the model for each dataset, the original 3-channel (RGB) images were used. The training pipeline is similar to the one used in \cite {kalinov2020warevision}: during the training, we randomly applied one of the following augmentation methods (flip, rotation, contrast, and brightness change; cropping, gamma changing, and channel shuffling) was applied to each image of the trained dataset in every epoch. The possibilities of each augmentation method were equal, and the total number of epochs was 1000.

\begin{table}[!t]
\caption{Comparison of Trained Neural Networks Metrics for Different Methods of Dataset Labeling}
\begin{center}
    \begin{tabular}{c|c|c|c}
                                 & Manual & Auto ArUco & Auto UR \\ \hline
    IoU train &      88.32  &    85.12       &      92.35         \\
    IoU validation  &     87.52   &    85.02        &     91.87          \\
    Presicion  &    89.97        &    86.46        &          94.81                  
    \end{tabular}
    
    \label{tab:compare2}
    \end{center}
\end{table}

After training, the data was tested on a separately prepared stand with other barcodes. At the same time, the lighting was fixed to neutral white and did not change. It was necessary to recreate the real situation of data detection. Automated data labeling was performed using the method that showed the least number of pixel errors in the image. The following metrics were chosen as criteria for assessing the network performance: 

\begin{itemize}
    \item \textit{Intersection over Union (IoU)} is the area of overlap between the predicted and ground-truth mask. It used during neural network training to estimate current state of recognition quality for \textit{train} and \textit{validation} data.
    \item \textit{Precision} is used to calculate the percentage of correct positive predictions among all predictions made on test data.
    
\end{itemize}

\begin{figure}[h]
%\vspace{-1em}
    \centerline{\includegraphics[width = 0.5\textwidth]{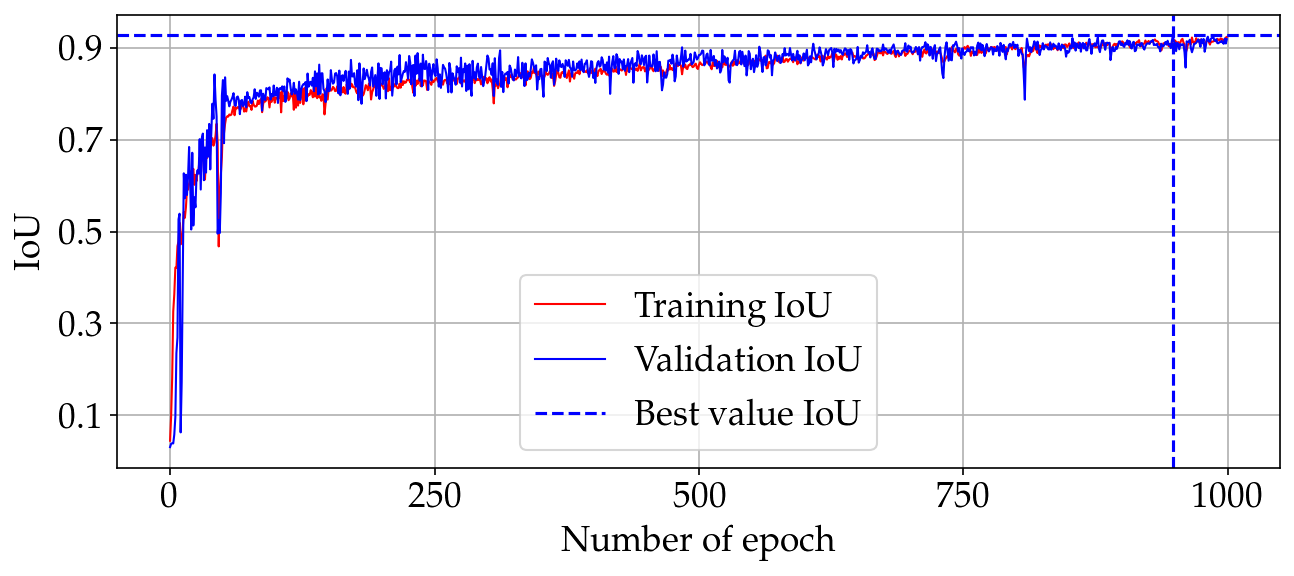}}
    \vspace{-1em}
    \caption{Training process of modified neural network architecture with DeepScanner approach.} %from \cite{kalinov2020warevision} 
    \label{fig:trainig_unet}
    \vspace{-1em}
\end{figure}

The comparison results of the three approaches for the selected metrics are presented in Table \ref{tab:compare2}. The results obtained using the proposed approach demonstrate high values of IoU and Precision. Precision of 94.81 in comparison with the indicators (91.56) obtained earlier in \cite{kalinov2020warevision} with minimal changes in the neural network architecture, is an indicator of the increase in accuracy due to the new data collected using our automated approach. The graph of training on the most efficient approach is presented in Fig. \ref{fig:trainig_unet}.

\section{Conclusions and Future Work}

We have developed DeepScanner, a robotic system for automated data labeling of 2D objects in the image. The system consists of a 6-DoF articulated robot, a camera and sensors that allow to capture images and automatically label data on them. Barcodes were selected as target objects for labeling. 

Our approach of data labeling significantly increases the dataset size. It also enhanced the speed of data labeling 240 times, and the accuracy of labeling compared to manual labeling 13 times. At the same time, this approach completely eliminates the possibility of errors in the labeling of the object itself (if the accurate labeling was made on the original frame).

We also trained CNN model. With the same architecture, the model obtained from automatically collected data using our approach had an increase in IoU on train and validation data, and the Precision metric on test data increased by 4.84\%, compared to the model trained on manually labeled data. Thus, the accuracy of the model when training on automatically obtained data is better than on data labeled manually.

The obtained results of the study can be used in future studies when designing a system for automated labeling of 3D objects based on an RGB-D camera. The use of an articulated collaborative robot will allow to automatically capture different sides of 3D objects, thus, improving the detection and segmentation accuracy.

\bibliographystyle{ieeetr} 
\bibliography{references}

\end{document}